\documentclass[conference]{IEEEtran}
\IEEEoverridecommandlockouts
\usepackage{cite}
\usepackage{amsmath,amssymb,amsfonts}
\usepackage{graphicx}
\usepackage{textcomp}
\usepackage[noend]{algpseudocode}
\usepackage{booktabs}
\usepackage[ruled,vlined]{algorithm2e}
\usepackage[table,xcdraw]{xcolor}
\usepackage{xcolor}
\def\BibTeX{{\rm B\kern-.05em{\sc i\kern-.025em b}\kern-.08em
    T\kern-.1667em\lower.7ex\hbox{E}\kern-.125emX}}
\begin{document}

\title{Parking Analytics Framework using Deep Learning
\thanks{This project is funded by RIOTU Lab in Prince Sultan University.}

\author{Bilel~Benjdira\textsuperscript{1}, 
        Anis~Koubaa\textsuperscript{1},
        Wadii Boulila\textsuperscript{1}, 
        Adel Ammar\textsuperscript{1} \\
$^1$RIOTU Lab, Prince Sultan University, Riyadh, Saudi Arabia.
}
}

\maketitle

\begin{abstract}
With the number of vehicles continuously increasing, parking monitoring and analysis are becoming a substantial feature of modern cities. In this study, we present a methodology to monitor car parking areas and to analyze their occupancy in real-time. The solution is based on a combination between image analysis and deep learning techniques. It incorporates four building blocks put inside a pipeline: vehicle detection, vehicle tracking, manual annotation of parking slots, and occupancy estimation using the Ray Tracing algorithm. The aim of this methodology is to optimize the use of parking areas and to reduce the time wasted by daily drivers to find the right parking slot for their cars. Also, it helps to better manage the space of the parking areas and to discover misuse cases.  A demonstration of the provided solution is shown in the following video link: https://www.youtube.com/watch?v=KbAt8zT14Tc. 
\end{abstract}


\section{Introduction}
Video Analytics using deep learning and artificial intelligence has been gaining increasing interest \cite{latif2021deep}. It has enabled several industrial applications ranging from surveillance \cite{Koubaa2020DeepBrain} and public safety \cite{Nasir2021, driftnet,melhim2018intelligent} to robotics perception \cite{Teixeira2021}, medical intervention \cite{Jabra2021COVID, spinal-cord}, and face recognition \cite{Koubaa2021Face,boulila2021deep}. The global market size of video analytics is valued at USD 5.9 billion in 2021 and is expected to reach USD 14.9 billion by 2026, as reported by Markets and Markets. 
Furthermore, the emergence of unmanned aerial vehicles (UAVs) has also enabled a vast array of applications (e.g., aerial surveys) for video analytics as it provides aerial views of the environment, allowing to collect aerial images and process them with deep learning algorithms \cite{Adel2021}. Parking analytics is one of these important applications in smart cities that leverages deep learning with UAVs to collect real-time data and analyze it to optimize parking revenues, improve parking resources allocations, and better manage the public space. In fact, parking coverage takes an average of 31\% of land use in big industrialized cities, and reaching 81\% in Los Angeles \cite{manville2005parking} for example.

Parking facilities management represents a significant challenge to authorities for urban planning. When a city becomes more populated, the number of vehicles increases, and parking slots become less available. Bad parking resource management leads to undesirable problems. It has been reported that 
searching for parking costs Americans \$73 Billion a year \cite{searching_for_parkings}. 
Naturally, this leads to traffic congestion, additional fuel consumption, higher risks of accidents, and more drivers' frustrations. Furthermore, in crowded areas such as shipping malls, drivers tend to park the closest possible to entry gates, which induces more load and traffic at specific parking slots than others. For this purpose, there is a need to develop parking analytics techniques that allow the analysis of the occupancy level of the parking slots for better management and operation.

In this paper, we consider the problem of real-time parking monitoring and data collection using UAVs, analyzing the parking occupancy over time using deep learning for better management and resource planning. This paper's contribution consists of developing an AI-based solution that combines YoloV3 object detection and DeepSort object tracker applied on aerial images of vehicles to analyze parking occupancy. The idea consists in processing the video stream using YoloV3 and DeepSort to generate a log file that provides statistical data about the parking occupancy in each batch of frames. The log file is then analyzed using data analytics tools to extract occupancy information of the parking area, including parking slot occupancy heatmaps, the number of occupied slots over time and the number of cars that occupied every slot.

The rest of this paper is organized as follows. In Section II, we discuss the recent works of parking analytics. In Section III, we present the proposed methodology of parking video analytics. In Section VI, we present the experimental results and performance evaluation of the proposed parking analytics solution. Finally, Section V concludes the paper, highlights the limitations of this work and future expected improvements.

\section{Related Works}
In the literature, several research works have been conducted on smart parking using video analytics or other approaches \cite{jemmali2021intelligent}. Khanna  et al. \cite{khanna2016iot} described an IoT solution for monitoring parking spaces. A mobile application acts as an interface between the IoT sensors and the cloud to book parking slots. The IoT module consists of ultrasonic sensors that are connected to a raspberry Pi. Nevertheless, no analytics or performance study were presented. Amato et al. \cite{amato2016car} proposed an approach for real-time parking status detection. The authors suggested classifying the parking space occupancy using CNN working on a Raspberry Pi camera. The developed system captures images of the parking and determines the occupancy status based on a trained CNN. These images are filtered by a mask that identifies spaces in the parking. The authors tested two CNN architectures, named mAlexNet and mLeNet, where m stands for mini. In \cite{bura2018edge}, Bura et al. designed a smart parking system using distributed cameras, edge computing, and DL algorithms. The authors developed a custom CNN model based on AlexNet neural network and having 1 input layer, 1 convolution layer, 1 ReLu, 1 max pooling, and 3 fully connected layers. The edge devices used in the proposed system are Raspberry Pi and Nvidia Jetson Tx2. Ke et al. \cite{ke2020smart} investigated the use of edge computing for smart parking surveillance and parking occupancy detection using real-time videos. The authors used two detection methods at the edge namely single shot multibox detector and background modeling detector. For the first one, the Mobilenet backbone network is used for the detection and TensorFlow Lite for the implementation. The background modeling detection is followed by blob detection, which transmits detected blobs to the server. In \cite{cai2019deep}, Cai et al. proposed a real-time video system. The authors combine deep convolutional neural networks and vehicle tracking filters to information across multiple image frames in a video sequence to remove noise generated by occlusions and detection failures. Cai et al. used Mask-RCNN for frame-wise instance segmentation and area-based threshold for parking identification. Jabbar et al. \cite{jabbar2021iot} developed an IoT Raspberry Pi-based parking management system to support university staff, students, and visitors to find available parking places. The developed system is composed of Raspberry Pi 4 B+ (RPi) embedded computer, Pi camera module, and GPS sensor. The authors suggested using ultrasonic sensors, LEDs, and Pi cameras to increase the detection of the occupancy of the parking spots.\\
Several smart parking management systems are already operational today in some large cities, such as LA Express Park \cite{LAExpressPark} and SFPark \cite{SFPark}. But they are generally based on the deployment of thousands of sensors in parking spaces, which is costly, prone to malfunction and transmission problems, and needs regular maintenance.

\section{Proposed Methodology}

\begin{figure}[!h]  
\begin{center}  
\includegraphics[width=9cm]{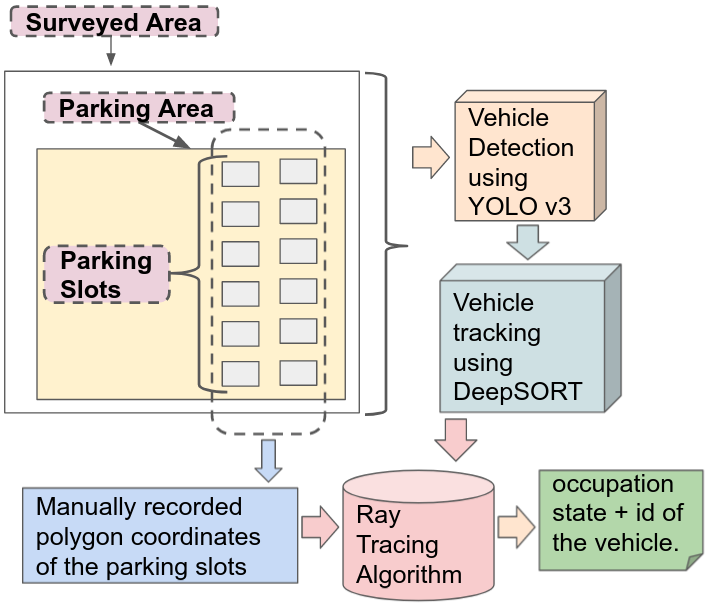}
\caption{\small \sl Flowchart of the methodology designed for Parking Areas Analysis
\label{fig:flowchart_of_the_method}}  
\end{center}  
\end{figure}

Figure \ref{fig:flowchart_of_the_method} depicts the proposed parking analytics approach. It is divided into three building modules (i.) The object detection module, (ii.) the object tracking module, and (iii.) the statistical log file module. 

The first module is the vehicle detection block using YOLOv3 \cite{YOLOv3} object detector. The second module is the vehicle tracking block using the Deep SORT algorithm \cite{deep_sort}. The third module includes the manual recording of the polygon coordinates of the parking slots, and  is running the ray tracing algorithm to verify if every tracked vehicle is included in any parking slot or not. All the data will be collected and recorded in a log file to generate the statistical insight of the parking. 

In the following subsections, we will describe the different modules in details and present the detailed algorithm of the method.

\subsection{Vehicle detection using YOLO v3}
The first step of our method is to detect all the vehicles on the video frame captured from a UAV or a surveillance camera. To ensure the best results on the statistics, the camera sensor should be perpendicular to the ground area of the parking. So that all the parking slots can be identified and the vehicles cannot be occluded by any other vehicle during the surveillance. Vehicle occlusion and unclarity of the parking slot boundaries are the most challenging conditions that we should avoid to ensure the correctness of the statistics. After fixing the viewpoint of the camera sensor to avoid these two conditions, we built a specific dataset for this purpose to train YOLO v3 \cite{YOLOv3} to detect five types of vehicles (Bus, Bicycle/Motorcycle, Truck, Pedestrian, Car). Any other object detection algorithm can replace YOLOv3 in this framework. YOLO v3 \cite{YOLOv3} begins by resizing the input images into a fixed size (e.g. \(416 \times 416 \)). This image is divided into a grid of cells. Every cell is of size \(32 \times 32 \). For example, if the size is \(416 \times 416 \), then the image will be divided into \(13 \times 13 \) cells. YOLO v3 uses one convolutional network to predict one tensor from the whole image. The tensor dimension is:
\begin{equation} \label{eq:6}
  (3*(5+C))  \times N \times N
\end{equation}
where:
\begin{itemize}
\item \( N\times N\): represents the  number of grid cells.
\item \(C\): represents the number of targeted classes.
\item \((3*(5+C))\): For every grid cell,  we detect five parameters: \( t_x, t_y, t_w, t_h \), and the box confidence score plus the probability for each one of the classes (\(C\)). This is done for every grid cell at three different scales to enhance the detector's performance on the big, medium, and small sizes of the objects. 
\end{itemize}
The images considered during our whole work are taken from a perpendicular view of the earth. Hence, we can take the hypothesis that the vehicles are circulating on the plane of the ground, which is of the equation:
\begin{equation} \label{eq:hyp_z_eq_0}
 [z = 0], (x, y, z) \in \Re 
\end{equation}
After applying YOLO v3 on the input images, we will get the set of vehicles inside it with the coordinates of the bounding box surrounding every vehicle. These coordinates will be easily converted into metric scale later. 
\subsection{Vehicle tracking using Deep SORT}
To identify every circulating vehicle inside the image, we need to use a multi-object tracking algorithm. We have chosen Deep SORT \cite{deep_sort}. This is a state-of-the-art algorithm that can be easily combined with any object detector to recognize the vehicle's identity through the video rapidly. For every frame, the Deep SORT algorithm takes all the detected vehicles by YOLO v3 and associates the proper ID to every one of them. If the vehicle is not previously seen in the scene, Deep SORT will associate a new ID. If the vehicle is previously seen in the scene, it will associate it to the ID of this tracked vehicle. The vehicle is represented using the following motion model state:
\begin{equation} \label{eq:9}
X= [x, y, h, r, \dot{x}, \dot{y}, \dot{h}, \dot{r}]
\end{equation}
where \(x\) and \(y\) represent the center of the bounding box of the vehicle, \(r\) is the aspect ratio of the vehicle, and \(s\) is its area. \(\dot{x}\) and \(\dot{y}\) represent the velocity of the vehicle. \(h\) is the bounding box height. Kalman filter is used in the image domain to predict the new motion state of the tracked vehicle based on its previous recorded motion states. To associate a newly detected object to the right ID, we calculate the Intersection over Union \(IoU\) between its bounding box and all the predicted box of the currently tracked vehicles. If \(IoU < IoU_{Min}\) for all the predictions, the association is rejected, and we give a new ID to the vehicle.  \par
Moreover, to reduce the frequency of identity switches and extended periods of object occlusion, Deep SORT uses a designed Convolutional Neural Network (CNN) to embed an appearance descriptor summarizing the object's visual features. To conclude the final association, we combine two metrics. The first is the Mahalanobis distance between the motion state of the detected object and the predicted motion state of every tracked vehicle. The second is the cosine distance between the appearance descriptor of the detected object and the appearance descriptor of every tracked vehicle in the database. A weight is used between the two metrics to estimate the final association. The last ID associated is used to identify every car entering the scene precisely. It is also used to analyze the behavior of every vehicle entering the scene separately. 

\subsection{Manual annotation of the polygon coordinates of the parking slot}
For every surveyed parking area, we should manually gather the polygon coordinates of every parking slot. If we have \(n\) parking slots, every parking slot \(PS_i , i =0..n\) will be considered during the processing as a series of points forming a closed polygon:
\begin{equation} \label{eq:k21}
PS_i = \{(x_0^{(i)}, y_0^{(i)}), (x_1^{(i)}, y_1^{(i)}),..., (x_{n_i}^{(i)}, y_{n_i}^{(i)})\} 
\end{equation}
To be a closed polygon, \(PS_i\) should meet the following constraint:
\begin{equation} \label{eq:k22}
(x_0^{(i)}, y_0^{(i)}) = (x_{n_i}^{(i)},..., y_{n_i}^{(i)})
\end{equation}
\subsection{Parking occupancy estimation using the Ray tracing algorithm}
For every frame, we check if every vehicles does belong to one of the \(PS_i , i =0..n\) parking slots. We use the approximation that if the central point of the vehicle \((x_c, y_c)\) is within the perimeter of a parking slot, we assign the vehicle to it. We loop this test over the whole list of the parking slots until the right assignment. \par
The problem of assessing if a specific point is within a closed polygon is coined in computational geometry as the  Point In Polygon problem (PIP). The Ray Tracing algorithm is one of the most used approaches to solve this problem. The idea behind it is to follow two steps. The first is to draw a ray starting from the selected point and oriented towards a random direction. Then, the second step is to count the number of intersections of this ray with the polygon's edges. If it is even, the point is outside the polygon. Otherwise, it is inside it. The whole procedure is formulated in Algorithm \ref{ray-tracing}. 
\begin{algorithm}[h]
\SetKwInOut{Input}{input}
\SetKwInOut{Output}{output}
\Input{$C(x_c, y_c)$: Center of the vehicle, ~\\ $Nb$: the number of parking slots, ~\\$PS_i$: the parking slots, \\$PS$: the set of parking slots,  }
\Output{$is\_inside$: Boolean value to assess if $C$ is within $PS_i$ or not, ~\\$OCC[]$: Parking Occupancy Array}
\BlankLine 
\SetAlgoLined 
$is\_inside$ = False ~\\ 
$n$ = 0 \# Number of Intersections ~\\ 
\ForEach{$i$ in $Nb$}
{
$OCC$[i] = 0 ~\\
}
\ForEach{$PS_i$ in $PS$}
{
\ForEach{$edge$ in $PS_i$}{~\\
\If{($ray\_intersects\_segment$($C$, $edge$)}
{
$n$ =  $n$ + 1 ; ~\\
}}
\If{($ODD$($number\_of\_intersections$)} 
{
$is\_inside$ = True ~\\
$OCC$[i] = 1 ~\\
continue 
}}
\caption{smart parking analysis for every vehicle in one frame}\label{ray-tracing}
\end{algorithm}
\begin{figure*}[!h]  
\begin{center}  
\includegraphics[width=16cm]{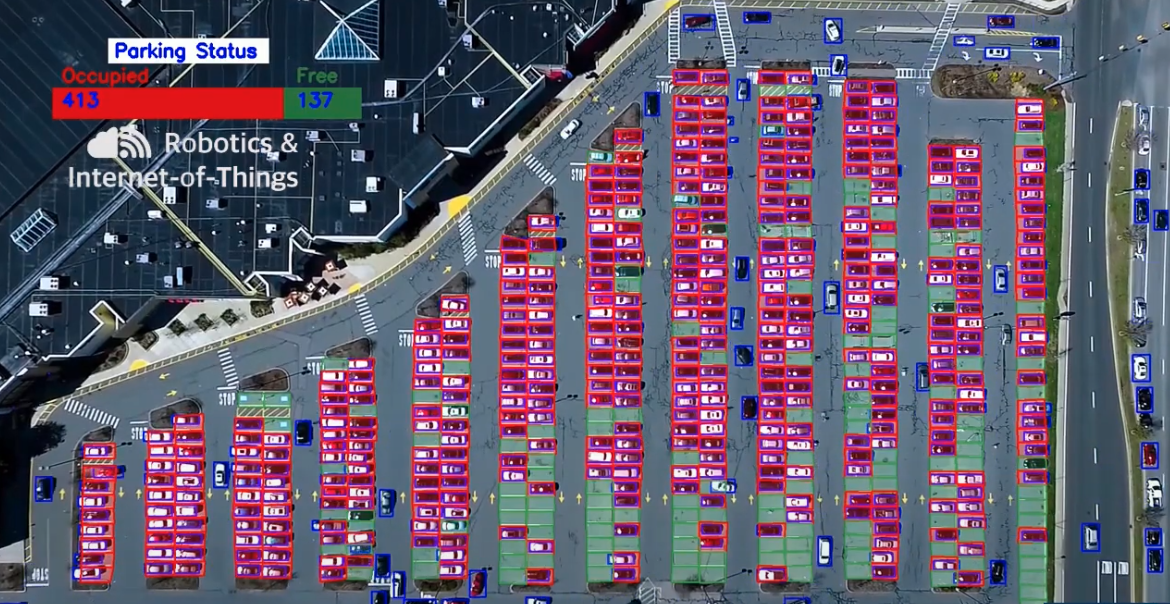}
\caption{\small \sl Parking Area Analysis in real time
\label{fig:parking_analysis}}  
\end{center}  
\end{figure*}

For every frame, the ray tracing algorithm is used inside a nested loop. We loop over all the detected bounding boxes of the vehicles. Then, for every bounding box, we loop over all the parking slots. If the value \(is\_inside\) is true, we stop testing for the rest of the parking slots. If the value is always false for all the parking slots, the vehicle is considered not associated with any parking slot. \par
On the other hand, an array of Boolean values is kept to store the occupancy of every tracing. 
The Algorithm \ref{ray-tracing} is run for every tracked vehicle in the frame. All the analysis procedure is based on it.

\section{Experimental Results}
Based on the proposed methodology described in the previous section, we will detail in this section some samples of the insights we can extract to better understand the parking area. In the following subsections, we will first describe the global insights generated by applying the framework to a sample parking surveillance video. Then we will describe deeper insights of the vehicles' movement inside the parking by showing figures and heatmaps that visualize better the occupancy of the parking. 
The machine we used during the inference has the following specifications:
\begin{itemize}
  \item CPU: Intel(R) Core(TM) i7-7700HQ CPU @ 2.80GHz
  \item GPU: NVIDIA GeForce GTX 1060 6 GB VRAM
  \item Memory: 32 GB
  \item CUDA version: 9.1
  \item Framework: Tensorflow 1.14 (with Keras API)
\end{itemize}
\subsection{Global insights generated using the framework}
We chose a sample video of a parking area with active movements of cars inside and outside the parking area. A screenshot describing the result of this analysis is made in Figure \ref{fig:parking_analysis}. For a complete video representation, we can refer to the following demonstration: https://www.youtube.com/watch?v=KbAt8zT14Tc. We can see in this demonstration that every vehicle circulating inside the area is detected using YOLO v3 and given an ID using Deep SORT. Then For every vehicle, we assign it to the right parking slot if possible. For every frame, we have a two-dimension array that describes for every parking slot ID two values. The first value is the occupancy status: occupied or not. It will show one if occupied and 0 otherwise. The second value is the ID of the vehicle occupying it. The value will be 0 if the slot is not occupied. For better visualization, we colored the parking slot in red if occupied, and green otherwise. We did not show the ID of the vehicle to not burden the final video visualization with many details. Only needed details are displayed. We added a status bar in the upfront to better estimate the occupancy status of the whole parking. The number of occupied slots is shown in the red region of the bar, and the number of unoccupied slots is shown in the green region of the bar. The real-time visualization of the parking status helps to better guide drivers to fast select the right slot without wasting time. In fact, locating unoccupied slots for the new coming vehicles remains one of the main problems that drivers face daily in modern cities. A screen can be installed in the entry of parking to guide newcomers to their right place in big parking areas.

\subsection{Deeper insights generated using the framework}
\begin{figure}[!h]  
\begin{center}  
\includegraphics[width=10.cm]{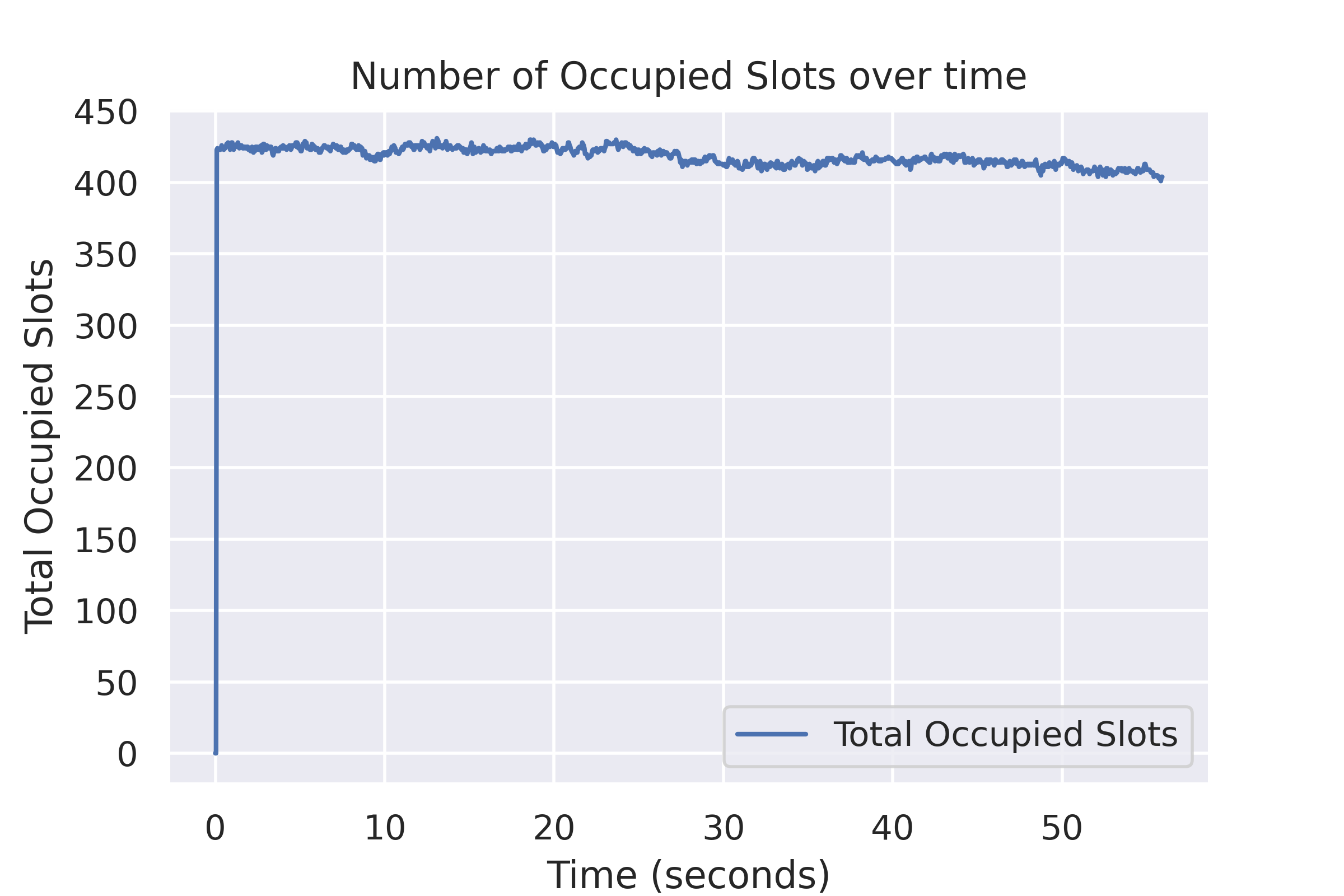}
\caption{\small \sl Number of vehicles existing in the whole parking area during a specific period of time. 
\label{fig:total_number_of_vehicles}}  
\end{center}  
\end{figure}

Although the previous video visualization of the parking shown in Figure \ref{fig:parking_analysis} gives us a clear visualization of the parking status in real-time, deeper insights can be extracted. For example, we can visualize the number of occupied slots over time to better study the parking occupancy during the time of day. A sample of this visualization is made for the video of Figure \ref{fig:parking_analysis} in Figure  \ref{fig:total_number_of_vehicles}. From this visualization, we can conclude how to better understand the sufficiency of the built slots for the daily needs of car parking. This represents a tangible measure to estimate if the parking area needs to be extended or redesigned.

\begin{figure}[!h]  
\begin{center}  
\includegraphics[width=10.25cm]{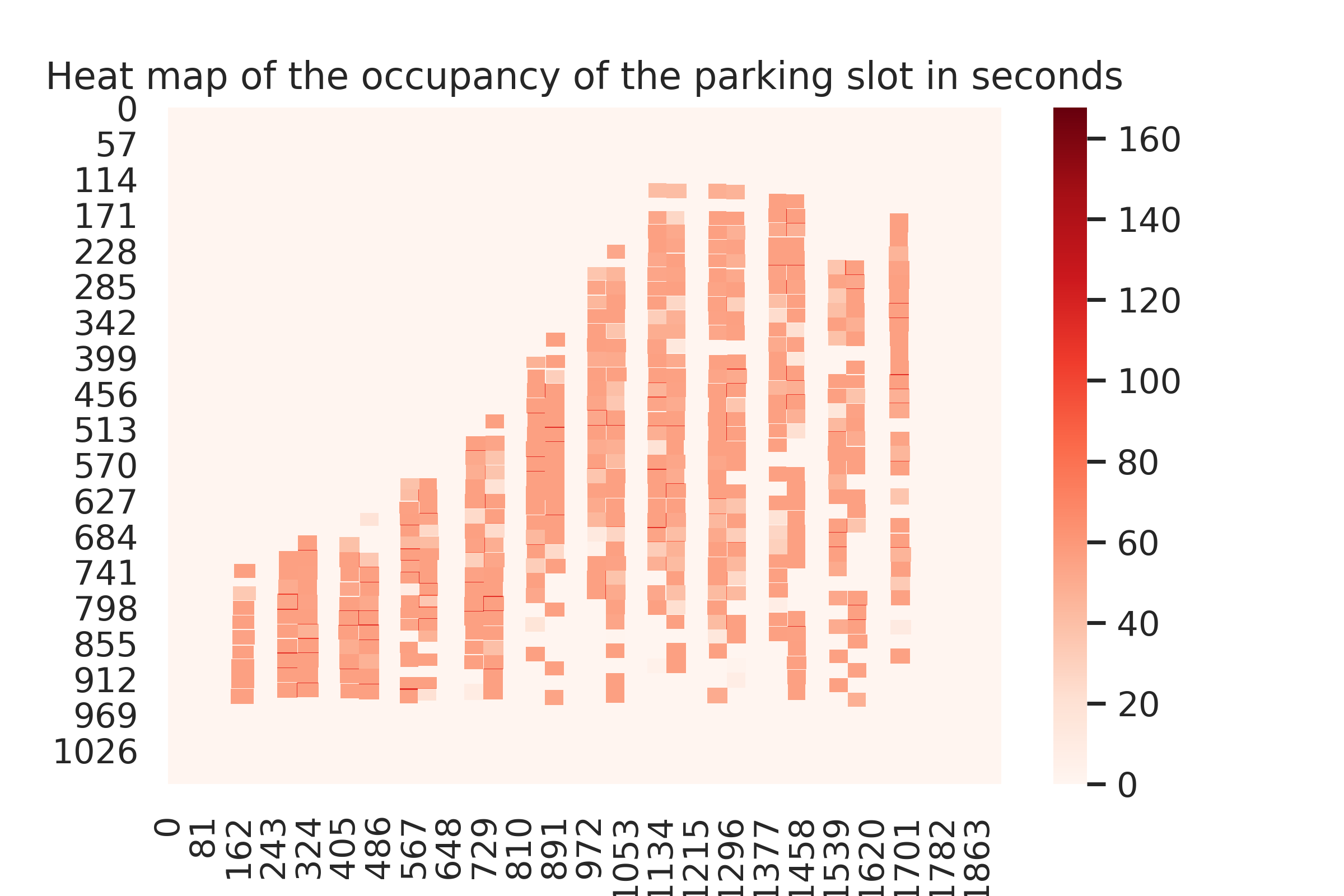}
\caption{\small \sl Heatmap representing the occupancy of every parking slot in seconds during a specific period of time. 
\label{fig:heatmap_occupancy}}  
\end{center}  
\end{figure}

We can get deeper beyond this visualization and understand the parking occupancy slot by slot. Here we need to choose another type of visualization, which is the heatmap. We took as an example here the parking occupancy of every slot in seconds. We extracted this for the video shown in Figure \ref{fig:parking_analysis} and draw it on the heatmap displayed on Figure \ref{fig:heatmap_occupancy}. We can see here that the color is changed following the duration of the slot occupation. This heatmap helps us to better manage the parking area by searching for the highly used slots and the rarely used slots in the parking. This helps the parking manager to make many decisions more confidently. For example, if some slots are always rarely used, we can reuse them for other purposes to better manage the space. Similarly, if all the slots are always highly occupied, the manager will be confidently encouraged for the extension of the parking or to redesign it to better use the provided space. 

\begin{figure}[!h]  
\begin{center}  
\includegraphics[width=10.5cm]{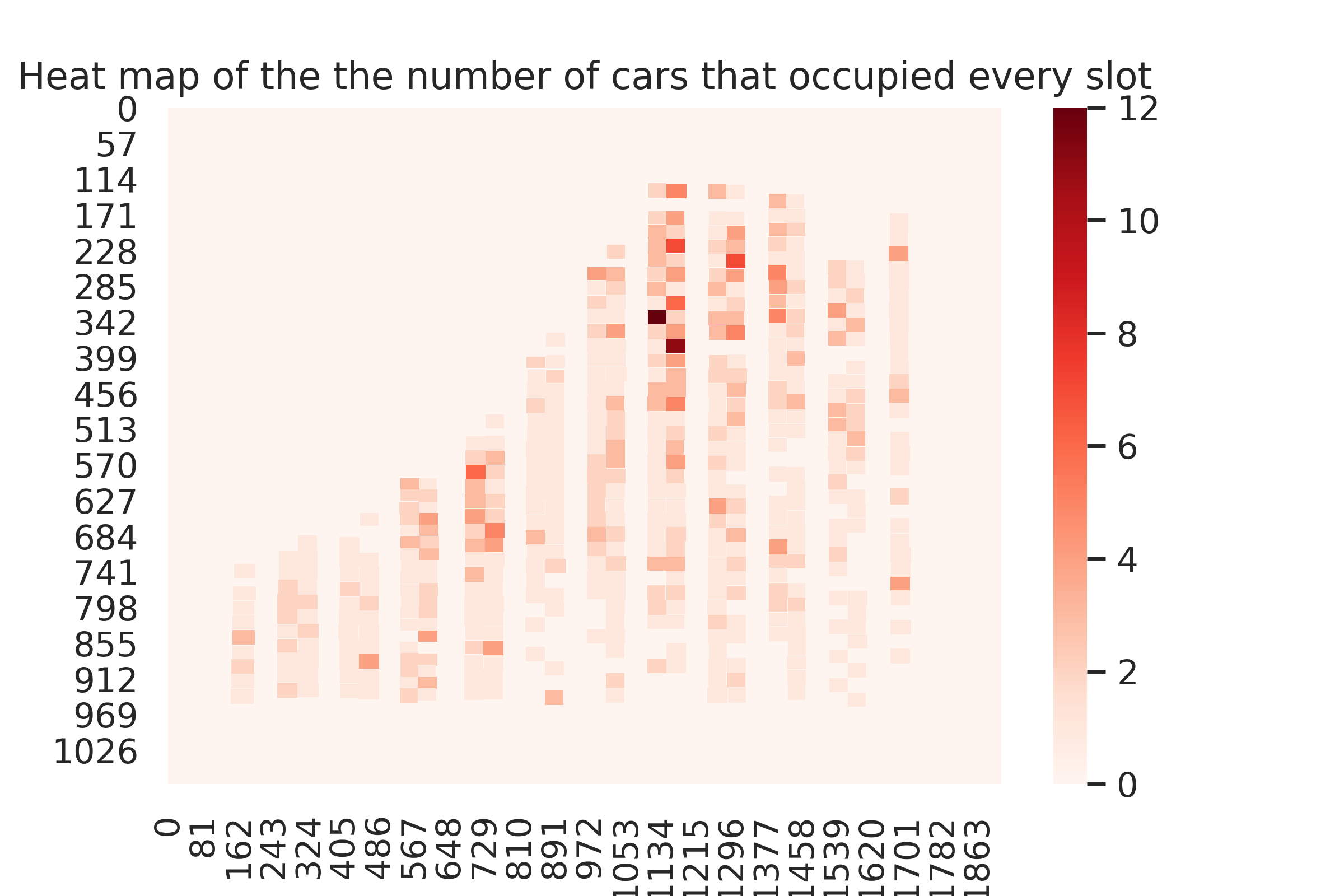}
\caption{\small \sl Heatmap representing the number of cars that occupied every parking slot during a specific period of time. 
\label{fig:heatmap_number_of_cars}}  
\end{center}  
\end{figure}

Another generated heatmap in our study is the number of cars that occupied every slot. We generated it here for the video of Figure \ref{fig:parking_analysis} and represented it in Figure \ref{fig:heatmap_number_of_cars}. We can see that most slots are occupied by one car during this specific period of time. 
This helps to understand accurately the behavior of the vehicles using the parking area. For example, this enables us to discover the cars that are left for a very long time in the parking. This represents a bad behavior that makes inflated occupancy of the parking beyond the real daily needs. Effective measures can be taken to only restrict parking usage for normal and daily needs. 

\section{Conclusion}
This study presented a methodology leveraging Deep Learning and Image Processing to better analyze and manage parking areas. We began by introducing the importance of this subject in modern cities. Then we summarized the research efforts that intersect with the matter presented. In the third section, we described the methodology and the four main building blocks constructing it. We formulated the pipeline and the algorithm used to extract occupancy statistics of the parking. Then, in the Experimental Results, we applied this pipeline on a sample video to show a real-time occupancy analysis of the parking. This makes a better visualization of the parking status and helps to better guide entering drivers to choose the right slot for their vehicles. Then, we extract more elaborated insights from the generated video, like curves and heatmaps.
To conclude, our solution is to solve one of the most recurring problems that drivers face daily in modern cities. Moreover, our solution is beneficial to solve many problems related to the management of parking areas: space management of the parking areas, avoiding misuse of the parking and long-lasting cars, design of the parking slots in the parking areas ... Nevertheless, our efforts need to be improved by working on randomly oriented camera viewpoints. This is a challenging problem because it primarily affects the accuracy of the extracted statistics. Also, the solution can be extended to target covered parking slots or partly occluded parking areas. Solving these two challenges helps better to adopt the solution on a more significant number of parking areas.

\bibliographystyle{ieeetr}
\bibliography{references}
\end{document}